\documentclass[journal]{IEEEtran}

\ifCLASSINFOpdf
\else
   \usepackage[dvips]{graphicx}
\fi
\usepackage{url}
\usepackage{graphicx}
\usepackage{stfloats}
\usepackage{bm}
\usepackage{amssymb,amsmath}
\usepackage{color}
\ifCLASSOPTIONcompsoc
    \usepackage[caption=false,font=normalsize,labelfont=sf,textfont=sf]{subfig}
\else
    \usepackage[caption=false,font=footnotesize]{subfig}
\fi

\begin{document}

\title{Adaptive Graph Convolutional Networks for Weakly Supervised Anomaly Detection in Videos}

\author{Congqi Cao, Xin Zhang, Shizhou Zhang, Peng Wang, and Yanning Zhang
\thanks{This work was partly supported by the National Natural Science Foundation of China (U19B2037, 61906155), the Young Talent Fund of Association for Science and Technology in Shaanxi, China (20220117), and the National Key R\&D Program of China (2020AAA0106900).}
\thanks{The authors are with the ASGO National Engineering Laboratory, School of Computer Science, Northwestern Polytechnical University, Xi’an 710129, China (e-mail: \{congqi.cao, szzhang, peng.wang, ynzhang\}@nwpu.edu.cn; zhangxin\_@mail.nwpu.edu.cn).}}

\markboth{Journal of \LaTeX\ Class Files, Vol. 14, No. 8, August 2015}
{Shell \MakeLowercase{\textit{et al.}}: Bare Demo of IEEEtran.cls for IEEE Journals}
\maketitle

\begin{abstract}
For weakly supervised anomaly detection, most existing work is limited to the problem of inadequate video representation due to the inability of modeling long-term contextual information. To solve this, we propose a novel weakly supervised adaptive graph convolutional network (WAGCN) to model the complex contextual relationship among video segments. By which, we fully consider the influence of other video segments on the current one when generating the anomaly probability score for each segment. Firstly, we combine the temporal consistency as well as feature similarity of video segments to construct a global graph, which makes full use of the association information among spatial-temporal features of anomalous events in videos. Secondly, we propose a graph learning layer in order to break the limitation of setting topology manually, which can extract graph adjacency matrix based on data adaptively and effectively. Extensive experiments on two public datasets (i.e., UCF-Crime dataset and ShanghaiTech dataset) demonstrate the effectiveness of our approach which achieves state-of-the-art performance.
\end{abstract}


\IEEEpeerreviewmaketitle

\section{Introduction}

\IEEEPARstart{W}{ith} the growing popularity of surveillance cameras, there is an urgent need to automatically detect and raise alarms for abnormal events \cite{xiao2015learning,sultani2018real,mestav2020universal}. Events in the real world are complex and diverse. Meanwhile, fine-grained annotation of training sets takes a lot of time and effort. Hence, weakly supervised anomaly detection (WSAD) becomes a popular research field. In previous work, WSAD has been formulated as a Multiple Instance Learning (MIL) task \cite{sultani2018real,zhang2019temporal,wan2020weakly,feng2021mist}. Sultani et al. \cite{sultani2018real} constructed a large-scale anomaly dataset and proposed a deep MIL ranking model to detect anomalies. Wan et al. \cite{wan2020weakly} improved this framework by replacing the max anomaly score selection policy with a $k$-max value selection policy. Yu et al. \cite{yu2021cross} proposed an effective cross-epoch learning strategy to introduce additional information from previous training epochs. However, the above methods ignore the spatial-temporal connection among video segments.

In recent years, several works have applied graph convolutional networks (GCNs) \cite{wang2018videos,shi2019two,lin2021multi,feng2020relation} over video sequence to model relations among different segments and learn powerful video representation. Zhong et al. \cite{zhong2019graph} used GCN to denoise normal segments in anomalous videos, and trained an action classifier with the obtained pseudo-labels. However, during the testing phase, the model only used current information despite capturing long-range temporal dependencies of the full video in the training phase. And the denoising process may clean up the anomalies, resulting in information loss. Wu et al. \cite{wu2020not} proposed a GCN with three parallel branches capturing long-range dependencies, local positional relation and the proximity of the predicted scores to describe different relationships among video segments respectively. However, three independent branches cannot model complex relationships coupled together in the video effectively and lead to slow iterative optimization. In addition, there are no learnable parameters in the adjacency matrix of the graph, hence the defined graphs may not be suitable for specific anomaly detection tasks.


\begin{figure*}[htp]
	\centering
	\includegraphics[width=0.85\linewidth]{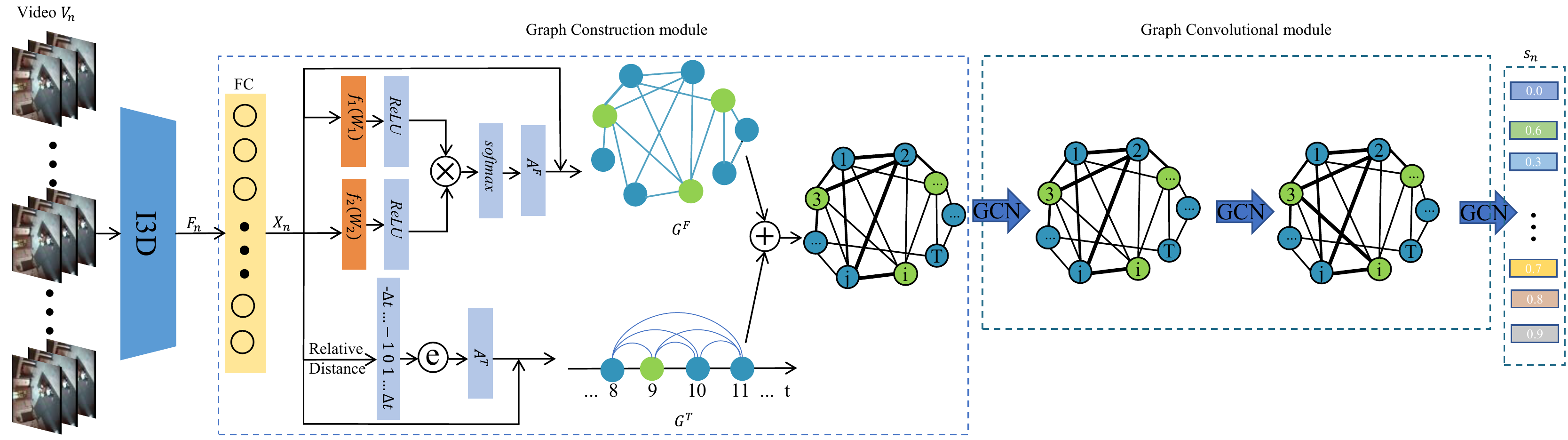}
	\caption{
Overview of our proposed method. First, the feature extractor (I3D) is used to extract segment features. Then, the Graph Construction module constructs a global graph to capture the long and short-range temporal dependencies among segment features. After that, the Graph Convolutional module further processes features with the learned graph and generates the anomaly scores at the segment-level.}
    \label{fig:framework}
\end{figure*}

In summary, we observe the following challenges in using GCN to model temporal contextual information for WSAD: (1) Only one network structure is used to model one type of relationship, or multiple independent branches are used to model different types of relationships separately, which cannot model multiple relationships coupled together within a sequence effectively. (2) Existing GCN methods ignore the fact that predefined graph structures are not optimal and it could be updated during the training process. In this paper, we propose an adaptive GCN to overcome these challenges. Our framework is shown in Fig. 1. For challenge 1, we use feature similarity graph to express the intrinsic association of segments that share similar characteristics. And we use temporal consistency graph to describe the intrinsic association of segments in a smooth undulating interval. However, both single feature similarity graph and single temporal consistency graph are not optimal to anomalous event detection. In order to make better use of the intrinsic correlations among segments, we integrate the two perspectives to jointly construct a global graph for spatial-temporal feature learning of events. For challenge 2, during the construction of the adjacency matrix, we break the limitation of manual setting. We consider the similarity among spatial-temporal features of video segments and other potential contextual semantic relationships to learn a video content-adaptive graph adjacency matrix that can be trained with the whole network simultaneously. As far as we know, we are the first to learn a global video content-adaptive graph adjacency matrix to model the contextual relationships among video segments for WSAD. Moreover, vanilla GCN suffers from over-smoothing problem, where the representations of all nodes converge to a smooth point, making them independent of the input features and causing gradient vanishing. We adopt to add residual connection \cite{he2016deep} in GCN to solve this problem for WSAD. In summary, our main contributions are as follows:

1) We propose to construct a global graph based on the similarity of spatial-temporal features and temporal proximity of video segments to better exploit the intrinsic correlation relationships among segments.

2) We propose to learn an adaptive adjacency matrix of the graph in an end-to-end manner to increase the flexibility of the model. And we propose an effective way to improve the learning ability of GCN by introducing residual connection.

3) Extensive experiments fully demonstrate the effectiveness of the proposed method, which achieves 84.67\% and 96.05\% frame-level AUC on UCF-Crime and ShanghaiTech datasets.

\section{Proposed Method}
\subsection{Feature Extraction}
For WSAD, the training set has \emph{N} training videos $\{V_{n}\}_{n=1}^{N}$ and corresponding weakly supervised labels \emph{\{Y$_{n}$\}$_{n=1}^{N}$}, where \emph{Y$_{n}$}$\in$\{0,1\}. Among which, \emph{{Y}$_{n}$}=1 indicates that \emph{V$_{n}$} contains at least one anomalous segment, but no temporal location annotation is provided. \emph{{Y}$_{n}$}=0 means that \emph{V$_{n}$} is entirely comprised of normal segments. Following the earlier MIL-based anomaly detection method \cite{sultani2018real}, each video \emph{V$_{n}$} is divided into \emph{T} non-overlapping segments which contains 16 frames before being fed to the feature extractor. The Inflated 3D (I3D) \cite{carreira2017quo} pretrained on the Kinetics dataset is used as the feature extraction network. The dimension of the extracted feature representation \emph{F$_{n}$} of \emph{V$_{n}$} is \emph{T}$\times$\emph{D}, where \emph{D} denotes the dimension of the segment features.


\subsection{Graph Construction Module}
Anomalous events occur in a continuous period of time and have a consistent behavior pattern over a period of time. Thus, in order to better model the temporal relationship of the video and better express the dynamic characteristics of the video, we combine the similarity of video segments’ spatial-temporal features and the degree of video segments’ temporal proximity to construct a global graph. In order to mitigate the curse of dimensionality, the output feature \emph{F$_{n}$} from the feature extractor first go through a fully connected layer to decrease the feature dimension to $T$$\times$$D'$, which is denoted as \emph{X$_{n}$}. For clarity, we omit the subscript $n$ in the following sections.

\subsubsection{Feature similarity graph}
We propose to construct a feature similarity graph \emph{$G^{F}$} with graph learning layer to represent the similarity as well as adjacency among segments in a video adaptively. Since we aim at capturing dynamic spatial-temporal dependencies, instead of defining the weights of two connected nodes based on the node inputs directly, we dynamically adjust them as the model is trained.
In detail, we first use two independent linear layers, i.e., \emph{f$_{1}$} and \emph{f$_{2}$}, followed with a ReLU activation, to embed the feature \emph{X}. The parameters of \emph{f$_{1}$} (and \emph{f$_{2}$}) are \emph{W$_{1}$} (and \emph{W$_{2}$}) with the size of $D'$$\times$$D^F$.
Then, we use dot product to measure the similarity as well as adjacency of any two segments in \emph{$G^{F}$}. Since an adjacency matrix should be non-negative, we bound the similarity to the range of $(0,1)$ with the normalized exponential function. Therefore, the adjacency matrix \emph{A$^F$} of \emph{G$^F$} is defined as:
\begin{equation}
	\ A^{F}={softmax}\left(relu \left(X W_{1}\right) relu \left(W_{2}^{T}X^{T}\right)\right)
\end{equation}
where \emph{A$^F$} is a $T$$\times$$T$-size learnable matrix representing the adjacency between segments based on similarity, which can be optimized  during training.

\subsubsection{Temporal consistency graph}
Video sequence has context association and temporal evolution characteristics between segments. In addition, the occurrence of events has a fluctuating duration time.
As pointed out in \cite{jayaraman2016slow,paul2018w,liu2022tcgl}, temporal consistency contributes to many video-based tasks. The temporal consistency graph \emph{G$^T$} is built directly based on the temporal structure of the video. Its adjacency matrix $A^{T} \in R^{T \times T}$ depends on the temporal position of the \emph{i}-th and \emph{j}-th segments:
\begin{equation}
	A_{i j}^{T}=\exp (-|i-j|)
\end{equation}
i.e., for the $j$-th segment, the closer it is to the $i$-th segment, the greater the weight assigned to their edge.

\subsection{Graph Convolutional Module}

We use $L$ GCN layers to fully exploit the relationship among segments. Specifically, for the $l$-th GCN layer, the graph convolution is implemented by:
\begin{equation}
	X^{l}=\sigma^{{l}}\left(\left(A^{F}+A^{T}\right)X^{{l}-1}W^{{l}}\right)+{f^l}(X^{{l}-1})
\end{equation}
where $X^{l-1}\in R^{T \times D^{l-1}}$ are the hidden features for all the segments at layer ${l}-1$, and {\emph{D$^{l-1}$}} is the dimension of the features. {$W^{l}\in R^{D^{l-1} \times D^{l}}$} is a trainable parametric matrix, and {$\sigma^l$} represents the activation function of the $l$-th layer.

In addition, we add residual connections between adjacent layers inspired by ResNet \cite{he2016deep} and use summation for aggregation, which is plug-and-play. If the dimension of the input channel is different from the number of output channels, a $1\times1$ convolution layer $f^l$ is inserted in the residual path to transform the input to match the output in the channel dimension. Otherwise, $f^l$ can be seen as an identity mapping.

\subsection{Training Loss of the Proposed Algorithm}

As mentioned before, each video only has a video-level label. For an abnormal video, we set the pseudo label $\left\{y_{i}\right\}_{i=1}^{T}$ of each segment to 1, while for a normal video, we set the pseudo label of each segment to 0.
Intuitively, segments with large abnormal scores in the abnormal video are more likely to be abnormal segments, while segments with large abnormal score in the normal segment are still normal segments. In order to expand the inter-class distance between abnormal and normal segments under weak supervision, we use $k$-max loss function \cite{wan2020weakly}. Specifically, given the abnormal score vector $s=\left\{s_{i}\right\}_{i=1}^{T}$ of video $V$ (where the instance index $n$ is omitted), we choose the top-$k$ elements in $s$ denoted as $S=\left\{s_{i}\right\}_{i=1}^{k}$, where $k=\left\lfloor\frac{{T}}{8}+1\right\rfloor$.
The final classification loss is the binary cross entropy between the predicted anomaly score and the pseudo label of the top-$k$ video segments:
\begin{equation}
	L_{k-\max }=-\frac{1}{k} \sum_{s_{i} \in S}\left[{y_{i}} \log \left(s_{i}\right)+(1-{y_{i}}) \log \left(1-s_{i}\right)\right]
\end{equation}

\section{Experiment Results}
\subsection{Datasets}

\subsubsection{UCF-Crime}

UCF-Crime is a large-scale dataset consisting of long untrimmed surveillance videos \cite{sultani2018real}. It covers 13 real-world anomalies. All videos are divided into two parts: the training set which consists of 800 normal and 810 anomalous videos, and the testing set which includes the remaining 150 normal and 140 anomalous videos.

\subsubsection{ShanghaiTech}

ShanghaiTech is a medium-sized dataset containing 437 videos collected from 13 scenes. To make it suitable for WSAD, Zhong et al. \cite{zhong2019graph} split the data into two subsets: the training set which is made up of 175 normal and 63 anomalous videos, and the testing set which contains 155 normal and 44 anomalous videos.

\subsection{Evaluation Metrics}

Following previous work, we use the frame-level receiver operating characteristic curve (ROC) and corresponding area under the curve (AUC@ROC) to evaluate the performance of our proposed method and comparison methods.

\subsection{Implementation Details}

In this work, 2048-D features are extracted from the “mix 5c” layer of I3D. The length of the videos varies widely, ranging from seconds to hours. Directly processing a very long video is not practical for batch-based training due to GPU memory constraints. Therefore, we uniformly extract $T$ segments at equal intervals to represent the entire video as \cite{zhong2019graph, wu2020not}. By default, we set $T$ to 150 for UCF-Crime and 100 for ShanghaiTech.
Note that fixed-length input is only used for training. As for inference, we use all the non-overlapping segments as others.
The FC layer for feature dimension reduction has 512 nodes. The output dimension of $f_1$ and $f_2$ are also 512. There are 3 GCN layers in total, which have 128, 32 and 1 node respectively. The first two layers are followed by a ReLU activation and a dropout function with a dropout rate of 0.6. The last layer is followed by a sigmoid activation function. Our method is trained end-to-end using the Adam optimizer with a weight decay of 0.0005 and a batch size of 64 for 100 epochs. The learning rate is set to 0.001. We use ten-crop augmentation in experiments as \cite{zhong2019graph}.

\subsection{Experimental Results and Discussions}

\begin{figure*}[!t]
	\centering
	\subfloat[Burglary]{\includegraphics[width=3cm]{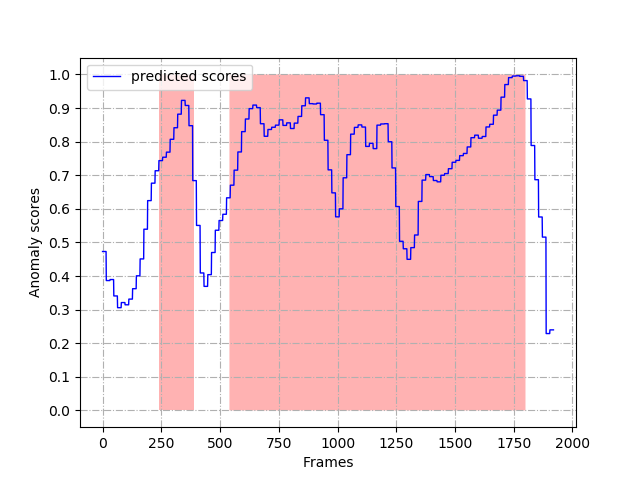}%
		\label{fig_first_case}}
	    \hspace{-0.6cm}
	\hfil
	\subfloat[Explosion]{\includegraphics[width=3cm]{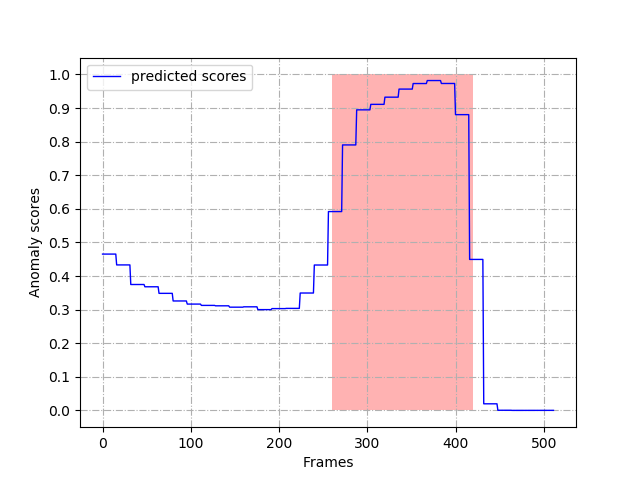}%
		\label{fig_second_case}}
	    \hspace{-0.6cm}
	\hfil
	\subfloat[RoadAccidents]{
		\includegraphics[width=3cm]{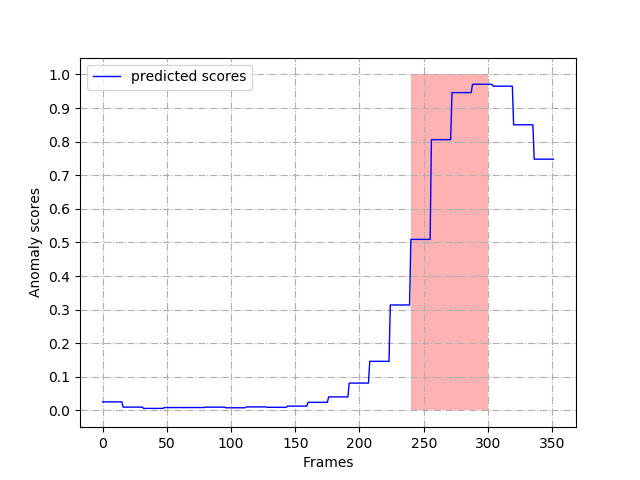}%
		\label{fig_third_case}}
	    \hspace{-0.6cm}
	\hfil
	\subfloat[Shooting]{\includegraphics[width=3cm]{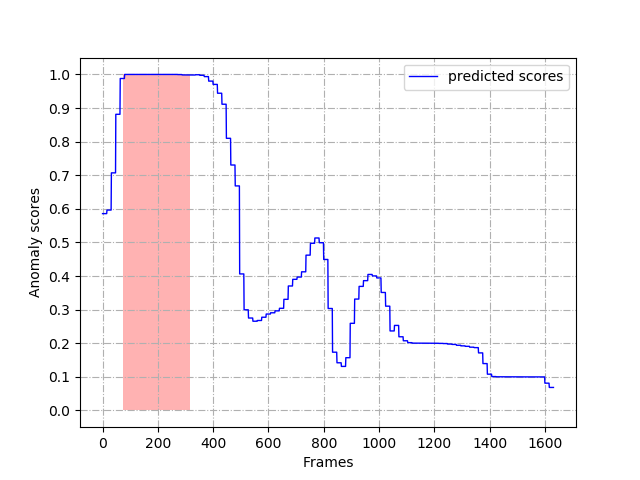}%
		\label{fig_forth_case}}
	    \hspace{-0.6cm}
	\hfil
	\subfloat[Vandalism]{\includegraphics[width=3cm]{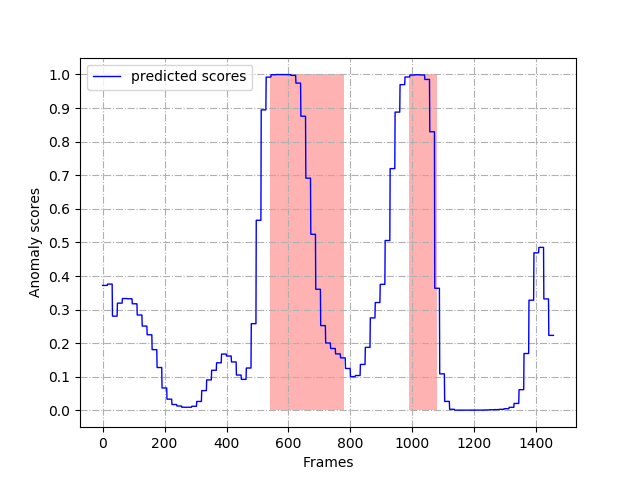}%
		\label{fig_fifth_case}}
	    \hspace{-0.6cm}
	\hfil
	\subfloat[Normal]{\includegraphics[width=3cm]{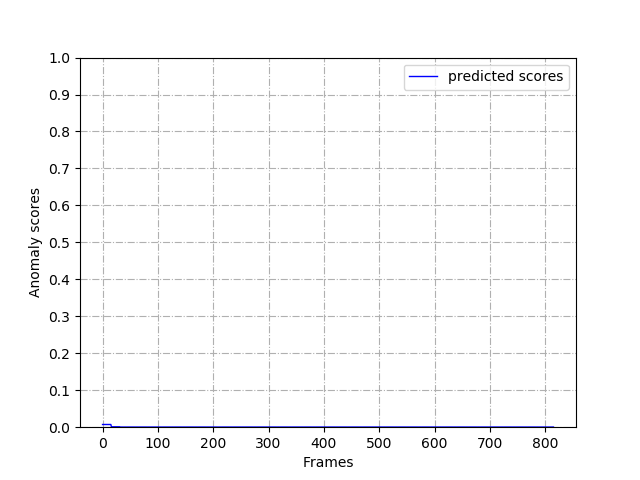}%
		\label{fig_sixth_case}}
	\caption{Visualization of the testing results on UCF-Crime. The red blocks in the graphs are temporal ground truths of anomalous events.}
	\label{fig:visualization}
\end{figure*}

We compare our method with the existing state-of-the-art methods. The results are shown in Table \ref{table:SOTA}. On UCF-Crime, our method outperforms MIST \cite{feng2021mist} by 2.37\%, XELM \cite{yu2021cross} by 2.52\%, and BN-SVP \cite{sapkota2022bayesian} by 1.28\%. Noticeably, using the same I3D-RGB features, our method shows a significant improvement over the previous GCN-based methods, outperforming Zhong et al. \cite{zhong2019graph} by 3.59\% and Wu et al. \cite{wu2020not} by 2.23\%. Furthermore, our method even outperforms the supervised method \cite{liu2019exploring} by 2.67\%, which added temporal and spatial labels to the UCF-Crime dataset and trained Convolutional 3D Network \cite{tran2015learning} with Non-local Network (NLN) \cite{wang2018non} for anomaly detection. On ShanghaiTech dataset, our method achieves better performance compared to previous weakly supervised methods \cite{sultani2018real,zaheer2020self,zaheer2020claws}. Especially, our method outperforms GCN-based weakly supervised method \cite{zhong2019graph} by a significant margin of 19.61\%, which indicates that our GCN module can capture the temporal dependence more effectively. Our method also improves the performance over AR-Net \cite{wan2020weakly} by 4.81\% even though it used additional multimodal features. To further illustrate the performance of our model, we visualize the temporal predictions of the model. As shown in Fig. \ref{fig:visualization}, our model is able to respond to both normal and abnormal events accurately.

Note there are other methods which achieve comparable or even higher performance. RTFM \cite{tian2021weakly} enforced large margins between segment features and utilized dilated convolutions and temporal self-attention for sequence modeling, which achieved 84.03\% and 97.21\% on the two datasets. However, without feature magnitude learning, its performance is 2.55\% and 3.73\% lower than our method, demonstrating that our method has the advantage of effective sequence modeling.
MSL \cite{li2022self} proposed a transformer-based Multi-Sequence Learning network (MSL). With VideoSwin features, its performance is 85.62\% and 97.32\% on the two datasets. However, compared with our method, which only needs 100 epochs for training, the two-stage MSL needs 500 epochs for optimization. Besides, our method has lower computation complexity and fewer learnable parameters.

\begin{table}[!tb]
	\begin{center}
		\caption{
			AUC(\%) performance comparison. The results with † are re-implemented with I3D features.
		}
		\label{table:SOTA}
		\setlength{\tabcolsep}{3pt}
		\begin{tabular}{l|c|c|c|c}
			\hline
			Method & Source & Feature & UCF-Crime & ShanghaiTech\\
			\hline
			Sultani et al. \cite{sultani2018real} & CVPR18 & C3D RGB & 75.41 & 83.17\\
			Sultani et al. \cite{sultani2018real}† & CVPR18 & I3D RGB & 76.92 & 85.32\\
			Liu et al. \cite{liu2019exploring} & MM19 & C3D RGB & 70.10 & /\\
			Liu et al. \cite{liu2019exploring} & MM19 & NLN RGB & 82.00 & /\\
			Zhong et al. \cite{zhong2019graph} & CVPR19 & C3D RGB & 81.08 & 76.44\\
			AR\_Net \cite{wan2020weakly} & ICME20 & I3D RGB+Flow & / & 91.24\\
            AR\_Net \cite{wan2020weakly}† & ICME20 & I3D RGB & 78.96 & 85.38\\
			SRF \cite{zaheer2020self} & SPL20 & I3D RGB & 79.54 & 84.16\\
			Wu et al. \cite{wu2020not} & ECCV20 & I3D RGB & 82.44 &/\\
			CLAWS \cite{zaheer2020claws} & ECCV20 & C3D RGB & 83.03 & 89.67\\
			MIST \cite{feng2021mist} & CVPR21 & I3D RGB & 82.30 & 94.83\\
			XELM \cite{yu2021cross} & SPL21 & I3D RGB & 82.15 & 87.83\\
			BN-SVP \cite{sapkota2022bayesian} & CVPR22 & I3D RGB & 83.39 & 96.00\\
			\hline
			Ours & / & I3D RGB & $\bm{84.67}$ & $\bm{96.05}$\\
			\hline
		\end{tabular}
	\end{center}
\end{table}
\subsection{Ablation Study}

\begin{figure}
	\centerline{\includegraphics[width=5cm]{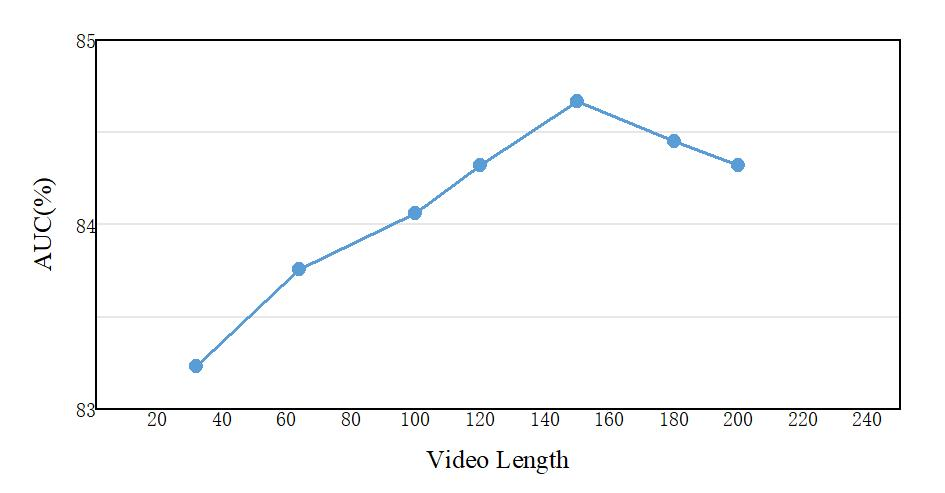}}
	\caption{Performance of different sampling lengths on UCF-Crime.}
    \label{fig:lengthT}
\end{figure}

To demonstrate the effect of sampling length, we vary the value of sampling length on UCF-Crime and plot the results in Fig. \ref{fig:lengthT}. The results show that with the increase of sampling length, the performance improves firstly and then decreases slightly. We observe the same trend on ShanghaiTech. It is probably because that too sparse sampling will lead to information loss, while too dense sampling will cause modeling difficulty.
We choose $T$=150 as the default setting for UCF-Crime and $T$=100 for ShanghaiTech based on the experiment.

To verify the efficiency of our proposed graph construction method of the adjacency matrix $A^F$, we perform ablation experiments on different construction methods. Dyn-A1 that we used is shown in Equation 1. Motivated by \cite{yang2020spatial}, Dyn-A2 is shown in Equation 5. Para-A denotes that the adjacency matrix is a parametric matrix, which contains ${T}^2$ parameters. Csim-A is computed by the cosine similarity scores of segment features. Inspired by \cite{fernando2021anticipating}, Jsim-A is computed by the Jaccard similarity scores of segment features. According to Table \ref{table:AF}, our proposed dynamic construction of \emph{A$^F$} achieves the best performance. Constructing the adjacency matrix dynamically in different ways has little effect on the model performance, but both are better than the adjacency matrix that is constructed fixedly at the beginning. If the graph structure does not depend on the input features of nodes at all, the final result is inferior, probably because the fixed design about the graph structure is limitative in learning.
\begin{equation}
	A_{(i, j)}=\frac{e^{2}\left(x_{i}, x_{j}\right)}{\sum_{j=1}^{N} e^{2}\left(x_{i}, x_{j}\right)}, e\left(x_{i}, x_{j}\right)=\left(w x_{i}\right)^{T} w x_{j}
\end{equation}

To verify the efficiency of constructing a global graph, we construct feature similarity graph and temporal consistency graph to train two independent branches and evaluate the performance of late fusion by averaging the results. The performance comparison on UCF-Crime dataset is shown in Table \ref{table:A}. It can be observed that constructing a global graph is more capable of expressing the complex relationships coupled together among the segments.

As for the residual connection, our model can obtain a performance gain of 2.5\% on UCF-Crime and 1.74\% on ShanghaiTech, which demonstrates its effectiveness.
\begin{table}[!t]
	\begin{center}
		\caption{
			AUC comparison of different graph adjacency matrix A$^F$ construction methods on UCF-Crime.
		}
		\label{table:AF}
		\begin{tabular}{c|c|c|c|c|c}
		    \hline
			Dyn-A1 & Dyn-A2 & Para-A & Csim-A & Jsim-A & AUC (\%)\\
			\hline
			$\checkmark$ & & & & & 84.67\\
			& $\checkmark$ & & & & 84.38\\
			& & $\checkmark$ & & & 82.89\\
			& & & $\checkmark$ & & 83.68\\
			& & & & $\checkmark$ & 83.27\\
			\hline
		\end{tabular}
	\end{center}
\end{table}
\setlength{\tabcolsep}{4pt}
\begin{table}[!tb]
	\begin{center}
		\caption{
			AUC comparison of different graphs on UCF-Crime.
		}
		\label{table:A}
		\begin{tabular}{l|l}
			\hline
			Method$\qquad\qquad$& AUC (\%)\\
			\hline
			feature similarity graph & 83.78\\
			temporal consistency graph & 83.26\\
			late fusion & 83.92\\
			global graph & 84.67\\
			\hline
		\end{tabular}
	\end{center}
\end{table}
\setlength{\tabcolsep}{1.4pt}
\section{Conclusions}

In this work, we propose an adaptive graph convolutional network for video anomaly detection. The method constructs a global graph considering both feature similarity and temporal disparity. Moreover, a graph learning layer is introduced to construct connections among segments in a video adaptively, which can capture spatial-temporal relationships among video segments effectively and enhance current temporal features. Extensive experiments on two typical datasets show that the proposed method achieves a high level of performance for video anomaly detection.

\bibliographystyle{IEEEtran}
\bibliography{egbib}
	
\end{document}